\documentclass{article}

 \PassOptionsToPackage{round}{natbib}

\usepackage[preprint]{nips_2018}

\usepackage[utf8]{inputenc} 
\usepackage[T1]{fontenc}    
\usepackage{hyperref}       
\usepackage{url}            
\usepackage{booktabs}       
\usepackage{amsfonts}       
\usepackage{nicefrac}       
\usepackage{microtype}      
\usepackage{listings}       
\usepackage[ruled,vlined]{algorithm2e} 
\usepackage{graphicx}       
\usepackage{inconsolata}    

\makeatletter
\renewcommand{\@noticestring}{}
\makeatother

\title{NengoDL: Combining deep learning and neuromorphic modelling methods}

\author{
Daniel Rasmussen \\
Applied Brain Research Inc.\\
Waterloo, ON, Canada \\
\texttt{daniel.rasmussen@appliedbrainresearch.com}
}

\begin{document}

\maketitle

\begin{abstract}
NengoDL is a software framework designed to combine the strengths of neuromorphic modelling and deep learning.  NengoDL allows users to construct biologically detailed neural models, intermix those models with deep learning elements (such as convolutional networks), and then efficiently simulate those models in an easy-to-use, unified framework.  In addition, NengoDL allows users to apply deep learning training methods to optimize the parameters of biological neural models.  In this paper we present basic usage examples, benchmarking, and details on the key implementation elements of NengoDL.  More details can be found at \url{https://www.nengo.ai/nengo-dl}.
\end{abstract}

\section{Introduction}

Deep learning and neuromorphic modelling share many methodological similarities: at their core, they are concerned with how groups of neurons, communicating via connection weights, can carry out some function of interest.  By ``neuromorphic modelling'' we mean the construction of models that include increased levels of biological detail, in an effort to understand or recreate the functionality of the brain (this is on a continuum with deep learning, rather than a sharp distinction).  Despite these computational similarities, researchers in the respective fields tend to be isolated from one another.  We usually think of deep learning in terms of abstract nonlinear optimization problems, and practitioners are rarely concerned with applying those methods to the study of the brain (with exceptions, e.g. \citealt{Kriegeskorte2015,Yamins2016}).  Correspondingly, there is a perception among neural modellers that deep learning methods are limited to abstract applications of artificial neural networks, and not of great help to those interested in studying how the brain works \citep{Kay2017}.

One significant outcome of this divide is that the tools of the two fields have become quite isolated.  Deep learning researchers use, e.g., TensorFlow \citep{Abadi2016}, Theano \citep{Team2016}, Caffe \citep{Jia2014}, or Torch \citep{Collobert2011}, while neuromorphic modellers use, e.g., Nengo \citep{Bekolay2014}, Brian \citep{Stimberg2013}, NEST \citep{Gewaltig2007}, NEURON \citep{Hines1997}, or PyNN \citep{Davison2009}. There is very little overlap between the two groups of users.

Our aim with NengoDL is to provide a tool that brings these two worlds together.  We want to combine the robust neuromorphic modelling API of Nengo with the power of deep learning frameworks (specifically TensorFlow).  In particular, there are three key design goals of NengoDL:

\begin{itemize}
\item Allow users to construct neuromorphic models using Nengo and then optimize model parameters using deep learning methods
\item Improve the simulation speed of neuromorphic models
\item Make it easy to construct hybrid models (e.g., inserting convolutional layers into a neuromorphic model, or converting a deep learning network to use spiking neurons)
\end{itemize}

In Section~\ref{sec:background} we discuss the two tools that form the basis of NengoDL (i.e., Nengo and TensorFlow).  Section~\ref{sec:usage} explains the basic features of NengoDL, with usage examples.  Section~\ref{sec:implementation} dives into the underlying implementation of NengoDL.  Finally, Section~\ref{sec:results} presents some benchmarking data, as well as results from some more complicated examples.

All of the source code for NengoDL can be found at \url{https://github.com/nengo/nengo-dl}, and installation instructions, more examples, and API descriptions can be found in the documentation at \url{https://www.nengo.ai/nengo-dl}.

\subsection{Related work}

As mentioned, NengoDL sits at the intersection between deep learning and neuromorphic modelling tools, combining Nengo \citep{Bekolay2014} and TensorFlow \citep{Abadi2016}.  The relationship of those tools to their respective fields is better described in their own papers, so we do not attempt to reiterate that here.  With regards to NengoDL itself, we are not aware of any similar tools that have attempted to combine deep learning and neuromorphic modelling methods.  There has been work at the intersection of deep learning and neuromorphic modelling \citep[e.g.,][]{Esser2015,Hunsberger2016,Kriegeskorte2015,Yamins2016,Lee2016}, which often involves developing specific software or hardware implementations that combine methods from the two fields. However, none of these efforts have developed a general modelling tool for others to use.

The most closely related work is the ``SNN Toolbox'' \citep{Rueckauer2017}.  The goal of that tool is to take a network constructed and trained using one of the deep learning packages (e.g., Theano or Caffe), and convert it into a special form of spiking neural network that will be able to match the performance of the source network as closely as possible.  Although this is something a user can do in NengoDL (e.g., see Section~\ref{sec:spiking_mnist}), the scope of NengoDL is more general.  In addition to supporting deep learning style networks, with NengoDL we are able to construct neuromorphic models (including spiking neural models), and support the simulation and optimization of both types of model (or mixtures of the two) in a unified framework.

\section{Background}
\label{sec:background}

In this section we give a brief introduction to the two tools we bring together in this work, Nengo \citep{Bekolay2014} and TensorFlow \citep{Abadi2016}.  Our goal is not to give a comprehensive or representative introduction to their features, but rather to focus on those elements that are most relevant to the upcoming discussion of NengoDL.  The interested reader can find more information at

\begin{itemize}
\item {\bf Nengo}: \url{https://www.nengo.ai}, \citet{Bekolay2014}
\item {\bf TensorFlow}: \url{https://www.tensorflow.org}, \citet{Abadi2016}
\end{itemize}

\subsection{Nengo}

Nengo is a software framework designed to enable the construction and simulation of large-scale neural models.  It has been used to develop, e.g., models of human motor control \citep{DeWolf2016}, visual attention \citep{Bobier2014}, inductive reasoning \citep{Rasmussen2014}, working memory \citep{Choo2010}, reinforcement learning \citep{Stewart2012,Rasmussen2017}, as well as large integrative models that combine these systems to produce end-to-end functional models of the brain \citep{Eliasmith2012a}.

There are a number of common characteristics of these kinds of models, which Nengo is designed to support:

\begin{itemize}
\item {\bf Temporal dynamics}: Nengo models are essentially temporal; they are simulated over time, and even a constant input (although inputs are rarely constant) will result in complex internal dynamics (the accumulation of neuron voltages, post-synaptic filtering, communication delays, online learning rules, etc.).
\item {\bf Complex neurons}: Nengo models typically use more complex neuron models than standard deep learning nonlinearities. In particular, these models are often simulated using spiking neurons.  Managing the parameters of these neuron models as well as their internal simulation is an important element of Nengo's design.
\item {\bf Complex network structure}: Neural models are often highly interconnected (e.g., containing large numbers of lateral and feedback connections), without the regular, feedforward, layer-based structure common in many deep learning models.  The Nengo model construction API has been designed to support this style of network.
\item {\bf Neuromorphic hardware}: There are a number of interesting neuromorphic hardware platforms that are in development or have been recently released \citep[e.g.,][]{Khan2008,Benjamin2014,Davies2018}.  Nengo's architecture has been designed to allow the same model to run across diverse hardware backends, with minimal cognitive load on the user.
\end{itemize}

Note that none of these issues are exclusive to neuromorphic modelling.  However, they are common or prominent concerns in that field, which has shaped the design emphasis of tools such as Nengo.  This is why it is important to combine the strengths of Nengo with the strengths of TensorFlow, rather than simply choosing one over the other.

\subsubsection{Architecture}

As mentioned, one of the important design goals of Nengo is to allow the same model to run transparently (from the user's perspective) across a wide range of hardware platforms.  Thus Nengo's architecture has been designed to provide a clean separation between the front-end code (which modellers use to define the structure/parameters of their network) and back-end implementation (which handles the simulation of that network on some underlying computational platform).  

\begin{figure}
\centering
\includegraphics[width=\textwidth]{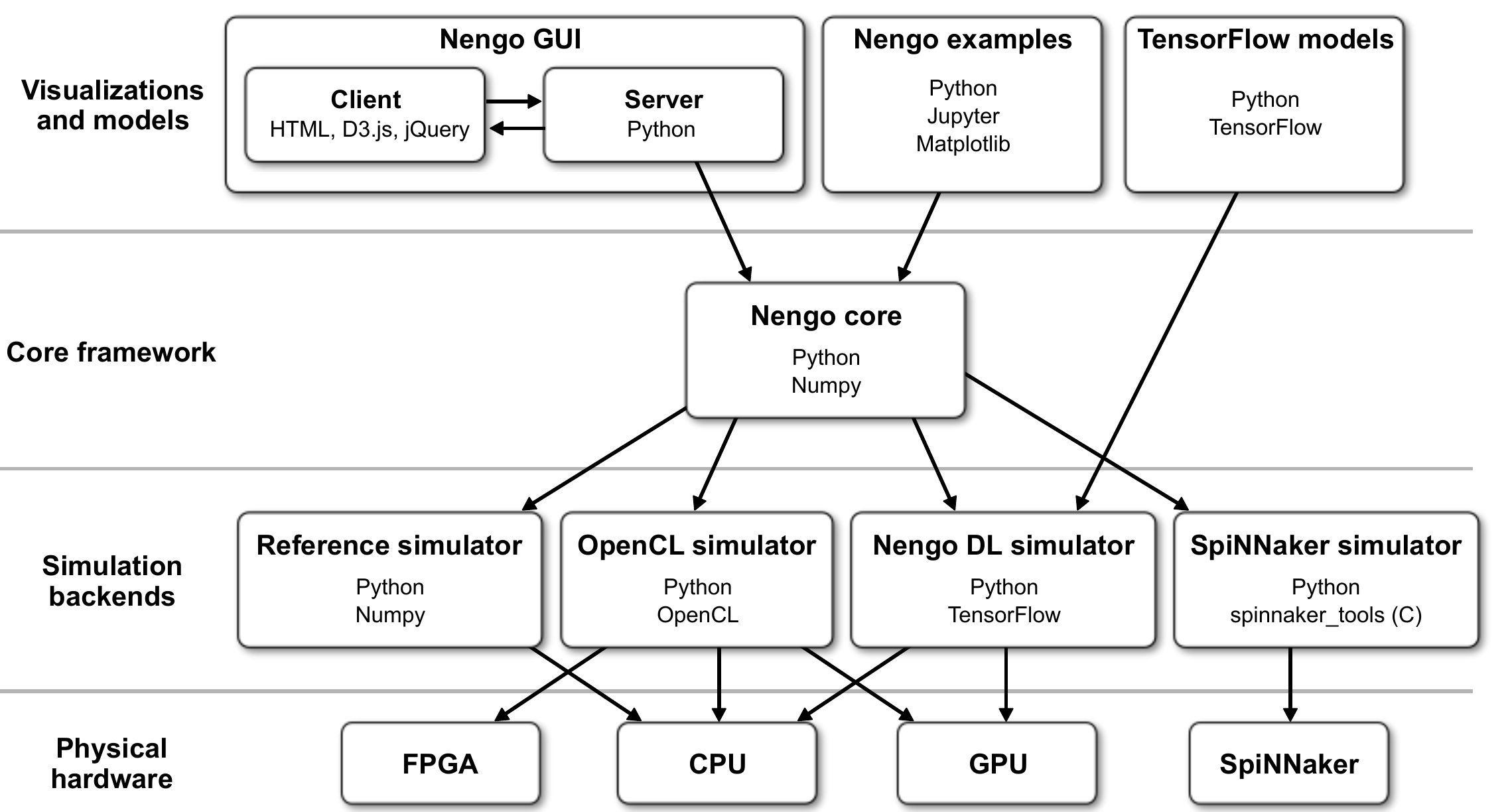}
\caption{Architecture of the Nengo ecosystem.  The primary role of NengoDL is as a Nengo simulator, meaning that it interfaces between the core Nengo modelling API and the underlying hardware.  However, NengoDL spans multiple levels, as it allows TensorFlow models to integrate in the same simulation environment, and also provides new user-facing functionality (augmenting the core framework).}
\label{fig:architecture}
\end{figure}

Users begin by constructing a \texttt{Network}, and populating it with the objects that define their model.  This is then passed to a \texttt{Simulator}, which encapsulates the back-end logic required to simulate that \texttt{Network}.  For example, the default \texttt{nengo.Simulator} will simulate a network on a conventional CPU, or the user can simply replace \texttt{nengo.Simulator} with \texttt{nengo\_ocl.Simulator} or \texttt{nengo\_spinnaker.Simulator} to run that same model on a GPU (using OpenCL) or SpiNNaker (custom neuromorphic hardware; \citealt{Khan2008}), respectively.  

In the case of NengoDL, we provide a \texttt{nengo\_dl.Simulator} that will simulate a Nengo network via TensorFlow (Figure~\ref{fig:architecture}).  Thus NengoDL resides primarily on the back-end side of the Nengo architecture (although it does provide some new user-facing features, which we discuss later).  In other words, the model construction process is largely unchanged from the user's perspective when switching from Nengo to NengoDL; any model constructed for the default \texttt{nengo.Simulator} will also run on \texttt{nengo\_dl.Simulator} and produce the same output (within the bounds of floating point precision).  Thus we give a brief introduction to the front-end side of Nengo here, but focus primarily on the back-end.  More details on the front-end can be found at \url{https://www.nengo.ai} or \citet{Bekolay2014}.

\subsubsection{Front-end objects}

A Nengo model is composed of 5 basic objects:

\begin{itemize}
\item {\bf Network}: Acts as a container for other Nengo objects
\item {\bf Ensemble}: A group of neurons
\item {\bf Node}: Used to insert signals into a model (e.g., sensory input)
\item {\bf Connection}: Used to connect Nodes or Ensembles, allowing objects to pass information to one another
\item {\bf Probe}: Extracts data from the model for analysis
\end{itemize}

We can think of these objects as defining a graph, where Nodes and Ensembles are the vertices and Connections are the edges.  Each of these objects, in addition to defining the structure of the graph, stores information about the parameters of that object (e.g., neuron types and bias values for the \texttt{Ensemble}, or synaptic filters and connection weights for the \texttt{Connection}).  Thus in the end we can think of a front-end Nengo network as a large data structure defining the structure and parameters of a model.  It is then the task of the back-end to take the information encoded in that data structure and construct an implementation that matches that specification.

\subsubsection{Back-end objects}

\begin{figure}
\centering
\includegraphics[width=\textwidth]{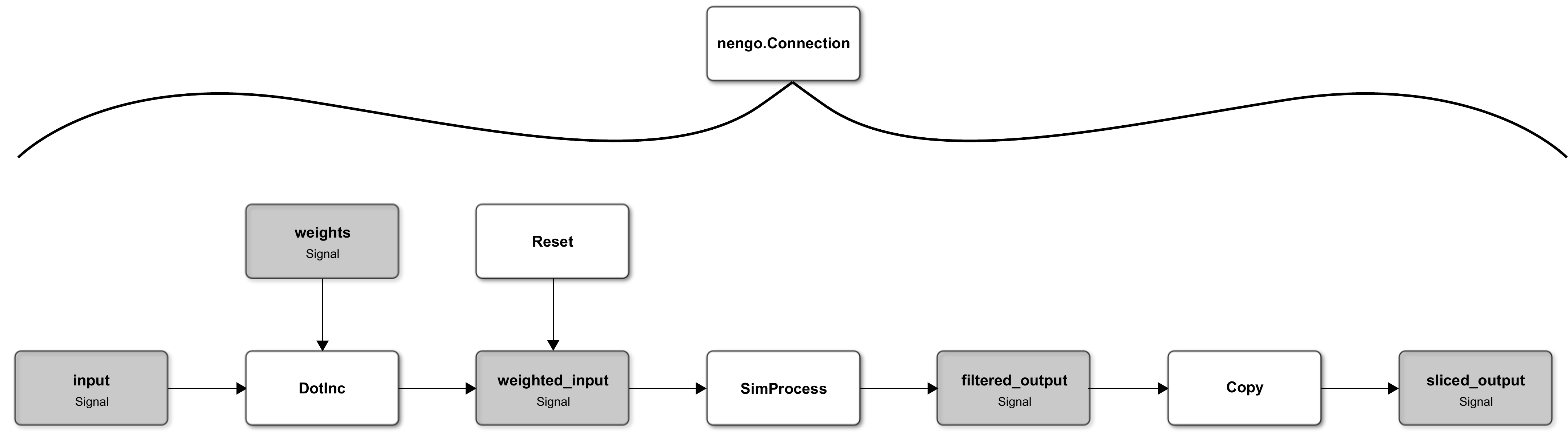}
\caption{An example of how a \texttt{Connection} is translated into lower level \texttt{Operations} and \texttt{Signals}.  The \texttt{DotInc} op multiplies the input signal (the source of the \texttt{Connection}) by the connection weights, and adds the result to another signal (which is \texttt{Reset} to zero at the beginning of each timestep).  Next a \texttt{SimProcess} op implements the synaptic filtering.  Finally, a \texttt{Copy} op copies the filtered input signal to the appropriate indices of the output signal (the destination of the \texttt{Connection}).}
\label{fig:connection_ops}
\end{figure}

In general, a Nengo back-end is free to process the input \texttt{Network} however it wants.  However, Nengo provides a builder that translates the high-level Nengo objects described above into a collection of lower-level operations (Figure~\ref{fig:connection_ops}).  This intermediate representation is often useful when designing a \texttt{Simulator}, as it is closer to the underlying computational operations that need to be executed on the back-end platform.  This intermediate representation consists of two basic objects:

\begin{itemize}
\item {\bf Signals}: A \texttt{Signal} is a generic tensor array that stores internal simulation values
\item {\bf Operators}: An \texttt{Operator} reads data from some number of input \texttt{Signals}, performs some computation, and writes the result to some output \texttt{Signals}
\end{itemize}

There are a number of basic \texttt{Operator} types, such as \texttt{Copy} (to copy a value from one signal to another), \texttt{ElementwiseInc} (computes the element-wise multiplication of two input signals and adds the result to some output signal), or \texttt{DotInc} (computes a matrix-vector multiplication).  There are also Operators for the different neuron types (e.g., \texttt{SimLIF}, which reads signals containing the input currents and internal state values of a group of leaky-integrate-and-fire neurons and computes output spikes) or online learning rules.

The task of the back-end is then to provide a concrete implementation for these operations.  For example, the default \texttt{nengo.Simulator} uses the Python \texttt{numpy} library, where \texttt{numpy.ndarray} is used to represent \texttt{Signals} and, e.g., \texttt{numpy.dot} is used to implement \texttt{DotInc}.  The first challenge for NengoDL is to do the same, but using TensorFlow to implement these basic operations.

\subsection{TensorFlow}

TensorFlow is a software framework developed by Google \citep{Abadi2016}.  Its primary use case is deep learning, but we can think of it more generally as a numeric computation library that we want to use to run a neural simulation.

TensorFlow uses a declarative programming approach, meaning that rather than directly specifying the steps of the program (imperative programming) the user specifies at a more abstract level the computations they want to perform.  This declarative programming looks a lot like the Nengo back-end framework described above; at its core it consists of \texttt{Tensors}, which represent values, and \texttt{Ops}, which perform computations on some input \texttt{Tensors} to produce output \texttt{Tensors}. The programmer begins with some input \texttt{Tensors}, and then builds up a computation graph by applying different \texttt{Ops} that represent various transformations.  For example, \texttt{y = tf.matmul(a, b)} adds a Tensor \texttt{y} to the graph that represents the matrix multiplication of two other tensors \texttt{a} and \texttt{b}.  The user can then ask TensorFlow to compute the value of \texttt{y} (or any other \texttt{Tensor} in the graph), and TensorFlow will translate that declarative specification into actual steps that are executed on the CPU or GPU to compute the value.

There are two key features of TensorFlow that we take advantage of in NengoDL:

\begin{itemize}
\item {\bf Automatic differentiation}: Specifying our programs via this declarative graph enables various automated manipulations of that graph.  For example, we can add an element to the graph that represents the derivative $\frac{\partial y}{\partial a}$, and TensorFlow will automatically add all the elements to the graph required to compute that derivative.  This makes it easy (from a user perspective) to apply gradient descent-based optimization methods; once we have specified the structure of our network in TensorFlow, we get the gradient calculations essentially for free.
\item {\bf Accelerator abstraction}: The term ``accelerator'' refers to any under-the-hood tool that allows a TensorFlow program to run faster.  The most common example is a GPU, but this can also include custom hardware or software optimizations.  The important feature from our perspective is that with the declarative programming style we do not need to worry about \emph{how} our program will run; once we have defined the structure of the computation, we can leave it up to TensorFlow to figure out how to best take advantage of the available accelerators to run that program.
\end{itemize}

To summarize, once we are able to accurately translate a Nengo model into a TensorFlow computation graph, we are able to automatically differentiate our Nengo models and get significant improvements in simulation speed.

\section{Using NengoDL}
\label{sec:usage}

We begin by describing the features and usage of NengoDL from a user perspective; Section~\ref{sec:implementation} goes into more detail on how NengoDL is implemented.  Our goal here is not to provide a manual for NengoDL; that purpose is better served by the documentation, available at \url{https://www.nengo.ai/nengo-dl}.  Instead we focus, at a relatively high level, on what users can do with NengoDL, in order to frame the upcoming implementation description.

\subsection{Running a model}
\label{sec:running}

The primary interface for NengoDL is through the \texttt{nengo\_dl.Simulator} class.  At its simplest, this is a drop-in replacement for the default \texttt{nengo.Simulator}.  A very simple model would look something like

\lstset{language=Python, numbers=left, columns=fixed, basicstyle=\ttfamily}
\begin{lstlisting}[firstline=2, lastline=13]
import nengo
import nengo_dl

with nengo.Network() as net:
    a = nengo.Node(output=[1])
    b = nengo.Ensemble(n_neurons=100, dimensions=1)
    nengo.Connection(a, b)
    p = nengo.Probe(b)

with nengo_dl.Simulator(net) as sim:
    sim.run(1.0)
    print(sim.data[p])
\end{lstlisting}

This creates a \texttt{Network} to hold our model (line 4), adds a \texttt{Node} that simply outputs a constant value of 1 (line 5), creates an \texttt{Ensemble} with 100 neurons and 1 represented dimension (line 6), connects the input \texttt{Node} to the \texttt{Ensemble} (line 7), and adds a probe to the output of the \texttt{Ensemble} (line 8).  Note that this is all front-end code, which is completely independent of the back-end being used.  We will not go into any detail on how to construct a Nengo model here; see \citet{Bekolay2014} or the documentation at \url{https://www.nengo.ai/nengo} for more information in that regard.

We specify the back-end by creating a \texttt{nengo\_dl.Simulator} (line 10).  We then run the simulation for one second (line 11) and print the data collected by the probe (line 12).  Although we are using NengoDL here, we could switch line 10 to \texttt{nengo.Simulator} and everything else would continue to function in the same way.

However, the NengoDL simulator also adds some new options for the user.  We can use \texttt{nengo\_dl.Simulator(net, device=`/cpu:0')} or \texttt{nengo\_dl.Simulator(net, device=`/gpu:0')} to run the model on the CPU or GPU, respectively.  Or we could use the \texttt{dtype=tf.float32/tf.float64} argument to control the floating point precision of the simulation.

The NengoDL simulator also has a \texttt{minibatch\_size} argument, which will configure the simulation to run multiple inputs in parallel.  That is,

\lstset{numbers=none}
\begin{lstlisting}[firstline=16, lastline=18]
with nengo_dl.Simulator(net, minibatch_size=10) as sim:
    sim.run(1.0)
    print(sim.data[p])
\end{lstlisting}

is functionally equivalent to

\begin{lstlisting}[firstline=21, lastline=25]
with nengo_dl.Simulator(net) as sim:
    for i in range(10):
        sim.run(1.0)
        print(sim.data[p])
        sim.reset()
\end{lstlisting}

The former will be much faster, as it takes better advantage of parallelism in the computations (see Section~\ref{sec:simulation_speed}).  

However, the output is not particularly interesting in this case, since the input is the same in all 10 instances (the constant input of 1 we specified when creating the input \texttt{Node}).  To take better advantage of batched simulations we need to use another new NengoDL feature, input feeds.

Input feeds allow the user to override the default value of any input \texttt{Node} in the model.  This is specified via the \texttt{data} argument of \texttt{sim.run}.  This takes a dictionary mapping \texttt{Nodes} to arrays, where each array contain the values we want that node to output at each time step.  For example, we could have the input node output a random number on each timestep, with different random numbers in each batch element, via

\begin{lstlisting}[firstline=28, lastline=32]
import numpy as np

with nengo_dl.Simulator(net, minibatch_size=10) as sim:
    sim.run(1.0, data={a: np.random.randn(10, 1000, 1)})
    print(sim.data[p])
\end{lstlisting}

Note the shape of the input array; the first dimension is the batch size (10), the second is the number of timesteps (1000, since we are running for one second with the default timestep of 0.001s), and the third is the output dimensionality of the node (1).

Again, this is not an exhaustive description of the features of the NengoDL simulator, see the documentation at \url{https://www.nengo.ai/nengo-dl/simulator} for more details and examples.  We hope here to convey the basic flavour of running models with NengoDL; that is, largely the same as working with the default Nengo simulator, but with a few extra bonuses.

\subsection{Training a model}
\label{sec:training}

An entirely new feature of NengoDL is the ability to optimize parameters of the model via deep learning training methods.  The default Nengo simulator also optimizes model parameters, but via a least squares optimization method \citep{Eliasmith2003}.  The advantage of this method is that it is fast and flexible (e.g., it does not require the model to be differentiable).  However, it can only optimize with respect to the inputs and outputs of a single layer, and is only applied to the output connection weights.  Deep learning methods allow us to jointly optimize across all the parameters in a model, allowing for more detailed fine-tuning.  Note that all the standard Nengo optimization methods are also available and used in NengoDL; we are simply adding a new set of optimization methods to our modelling tool set.

These methods are accessed via the new \texttt{sim.train} method.  For example, we could train our example network from above to compute the square function:\footnote{Note that this code is only intended to introduce the syntax; it would not result in particularly effective training if we were to run it.  Better performance would require a more complicated Nengo model, which we are trying to avoid in this description.  Various full functional examples can be found at \url{https://www.nengo.ai/nengo-dl/examples}.}

\lstset{numbers=left}
\begin{lstlisting}[firstline=35, lastline=46]
import tensorflow as tf

inputs = np.random.randn(50, 1000, 1)
targets = inputs**2

with nengo_dl.Simulator(net, minibatch_size=10) as sim:
    sim.train(
        data={a: inputs,
              p: targets},
        optimizer=tf.train.AdamOptimizer(),
        n_epochs=2,
        objective={p: nengo_dl.objectives.mse})
\end{lstlisting}

When performing this style of optimization we need to specify the input and target values (for each entry in the input array, we want the network to output the corresponding value from the target array).  In line 3 we create a random input array; this works much the same as the \texttt{data} example above, with axes corresponding to batch size, number of timesteps, and the dimensionality of the input node, respectively.  Note that the batch size is the total number of elements in the training data set; these will be split into chunks of \texttt{minibatch\_size} elements during training, and the training will run through all 50 items in the dataset \texttt{n\_epochs} times (line 11).  We pass the inputs to the \texttt{train} function as a dictionary that maps input nodes to input arrays (line 8), as we did with \texttt{data}.

Specifying targets works in much the same way, but with respect to output \texttt{Probes} instead of input \texttt{Nodes} (lines 4 and 9).  Semantically, this specifies that when the input node outputs the values from the \texttt{inputs} array, we expect to see the corresponding \texttt{targets} values at the output probe.  It is then the goal of the training process to optimize the parameters in between \texttt{a} and \texttt{p} so as to make that happen.

On line 10 we specify the optimization method that should be used during the training process.  TensorFlow provides a number of standard optimization methods, any of which can be used with NengoDL (or any custom optimizer that conforms to TensorFlow's optimizer API).  

Note that most deep learning optimization methods rely on some version of gradient descent, which means that the network needs to be differentiable.  In many neuromorphic models this is not the case (e.g., the spiking LIF neuron model is not differentiable), so applying these optimization methods restricts the kinds of models that can be studied.  However, in many cases we can achieve good performance by training with a rate-based approximation of the spiking neuron model (which is differentiable), and then using those same trained parameters during inference with the spiking neuron model \citep{Hunsberger2016}.  NengoDL will perform these transformations automatically (swapping between rate and spiking neurons for training and inference) for neuron types that have a differentiable rate-based approximation.  This allows users to build a spiking neuron model and then optimize it with gradient-descent based training methods, with all the underlying details handled transparently by NengoDL.  See Section~\ref{sec:spiking_mnist} for a demonstration of this idea in practice.

Finally, on line 12 we define the the objective function.  This is the function applied to the output of the given probe in order to generate an error value, which the optimizer will then seek to minimize.  Passing \texttt{nengo\_dl.objectives.mse} will use the common Mean Squared Error function, or the user can pass an arbitrary function that maps outputs and targets to an error value using TensorFlow operations.

More information on the features and usage of the \texttt{sim.train} function can be found at \url{https://www.nengo.ai/nengo-dl/training}.

\subsection{Inserting TensorFlow code}
\label{sec:tensornode}

Another key feature of NengoDL is the ability to combine deep learning-style networks with Nengo neuromorphic-style networks.  For example, we could use a state-of-the-art convolutional vision network to extract features from raw input images, and then connect the output of that network to a spiking neuromorphic model.  This gives us the best of both worlds, allowing us to choose whichever paradigm is most appropriate for different aspects of a model.

NengoDL translates a Nengo model into a TensorFlow graph, so once that process is complete the user can, if they want, add whatever additional elements they want by working directly with that TensorFlow graph.  However, the underlying TensorFlow graph can be quite complex, and it may not be obvious how to correctly insert code into that graph.  In addition, such an approach splits the model definition across two qualitatively different frameworks.  The key goal of NengoDL is to combine methodologies, so we would like a way to write TensorFlow code that smoothly integrates with the Nengo model definition.

This is accomplished through the \texttt{nengo\_dl.TensorNode} class.  \texttt{TensorNodes} are analogous to standard Nengo \texttt{Nodes}, except they integrate natively with TensorFlow code.  The user writes some TensorFlow code (or reuses an existing network) that maps some input \texttt{Tensor} to an output \texttt{Tensor}.  They then pass that as a function to a \texttt{TensorNode}, which encapsulates that code as a Nengo object.  The \texttt{TensorNode} can be added to a Nengo \texttt{Network} and connected to other parts of the model via \texttt{Connections}, the same as \texttt{Ensembles} and \texttt{Nodes}.  Any values received from input \texttt{Connections} to the \texttt{TensorNode} will be passed as inputs to the TensorFlow function, and the output values of that function will be passed along any outgoing \texttt{Connections}.  For example, we could add a \texttt{TensorNode} to our example network from Section~\ref{sec:running} that applies a dense weight layer to the signal from the input node \texttt{a}, and sends the resulting value to the ensemble \texttt{b}:

\begin{lstlisting}[firstline=49, lastline=54]
with net:
    def tensor_func(t, x):
        return tf.layers.dense(x, 100, activation=tf.nn.relu)
    t = nengo_dl.TensorNode(tensor_func, size_in=1)
    nengo.Connection(a, t)
    nengo.Connection(t, b.neurons)
\end{lstlisting}

First we define the TensorFlow function, which takes two input variables: the current simulation time, \texttt{t}, and the value from any incoming \texttt{Connections} to the \texttt{TensorNode}, \texttt{x} (line 2).  Then we apply whatever TensorFlow ops we would like in order to compute the \texttt{TensorNode} output; in this case we are applying the \texttt{tf.layers.dense} function with 100 output nodes, which will create a dense weight matrix and apply the \texttt{relu} nonlinearity to the output (line 3).  Next we create the \texttt{TensorNode}, passing it the function we defined and specifying the dimensionality of the function input \texttt{x} (line 4).  Finally we connect up the inputs (from node \texttt{a}) and outputs (connecting directly to the neurons of ensemble \texttt{b}) (lines 5-6).

NengoDL also provides the the \texttt{nengo\_dl.tensor\_layer} function, an alternate interface for creating \texttt{TensorNodes} designed to mimic the familiar layer-based syntax common to many deep learning packages.  This is simply ``syntactic sugar'' that combines the creation of a \texttt{TensorNode} and the \texttt{Connection} from some input object to that \texttt{TensorNode} in one step.  For example, we could redo the above example using \texttt{tensor\_layer}:

\begin{lstlisting}[firstline=57, lastline=60]
with net:
    t = nengo_dl.tensor_layer(a, tf.layers.dense, units=100,
                              activation=tf.nn.relu)
    nengo.Connection(t, b.neurons)
\end{lstlisting}

In these simple examples we could have easily achieved the same result using normal Nengo objects.  However, more complicated deep learning network architectures may not be easily expressed through the Nengo API, which is where the value of \texttt{TensorNodes} becomes more apparent.  See \url{https://www.nengo.ai/nengo-dl/examples/pretrained-model} for a more in-depth example.  More details on the features of \texttt{TensorNodes} can be found at \url{https://www.nengo.ai/nengo-dl/tensor-node}.

\section{Implementation}
\label{sec:implementation}

We now provide more detail on how NengoDL is implemented.  Knowledge of this infrastructure is not required to use NengoDL, but is helpful for advanced users who want to do something like add new NengoDL neuron models.  The implementation also showcases some somewhat esoteric uses of TensorFlow, which may be of interest to other TensorFlow users.

\subsection{TensorFlow mapping}

As discussed in Section~\ref{sec:background}, Nengo produces a back-end representation consisting of \texttt{Signals} and \texttt{Operators}, which we need to map into a TensorFlow computation graph.

\subsubsection{Signals}

A seemingly natural solution would be to map \texttt{Signals} to \texttt{Tensors}.  However, in Nengo we often have multiple \texttt{Operators} that all want to write to the same \texttt{Signal}, or parts of a \texttt{Signal}, and then other \texttt{Operators} that want to read the final result of those combined writes (rather than the output from any individual \texttt{Operator}).  \texttt{Tensors} do not naturally support this style of processing; once a \texttt{Tensor} has been created it cannot be modified, except by creating a new \texttt{Tensor}.  That is, if three \texttt{Operators} all increment the same output \texttt{Signal}, that will actually result in a new output \texttt{Tensor} each time.  Thus we need to think of \texttt{Signals} as a more abstract representation, where writing to the same \texttt{Signal} may represent writing to various different \texttt{Tensors}.

To manage this bookkeeping we use a data structure called \texttt{SignalDict}.  This manages the mapping between \texttt{Signals} and the \texttt{Tensor} representing the current value of that \texttt{Signal}.  For example, imagine we have a \texttt{Signal} $s$ with current value $x$.  Suppose an \texttt{Operator} wants to add 1 to the value of $s$.  This will result in a new value $y = x + 1$, which will then be stored in the \texttt{SignalDict} as the current value of $s$.  Then when another \texttt{Operator} wants to add 2 to the value of $s$ we look up the current value $y$, create a new value $z = y + 2$, and store that again as the new value of $s$.  Thus all the \texttt{Operators} have the illusion that they are reading and writing to the same signals, even though that \texttt{Signal} may actually be represented as an interconnected graph of \texttt{Tensors}.

A second issue alluded to above is that in Nengo we often want to write to some subset of the elements in a \texttt{Signal} array.  \texttt{Tensors} are not designed to support this kind of operation; it is not possible to modify parts of a \texttt{Tensor} in-place, we can only create entirely new \texttt{Tensors}.  It is possible to achieve similar effects using conditional TensorFlow operations, but this is slow and inefficient (for example, if we wanted to increment just one element in a 1000-dimensional vector $x$, we would have to create a new 1000-dimensional vector $y$ that is just a copy of $x$ in 999 of its elements).

Fortunately there is another TensorFlow data structure that does support in-place modification of elements: \texttt{Variables}.  \texttt{Variables} are usually used to represent things like connection weights, and because we want optimizers to be able to iteratively update those weights during the training process (without generating a new copy of all the model's parameters each time) they are designed to support in-place modification.  However, more generally we can just think of \texttt{Variables} as stateful \texttt{Tensors}, which is exactly what we want for our \texttt{Signal} values.  So in practice the \texttt{SignalDict} will actually maintain a mapping from \texttt{Signals} to \texttt{Variables}, so every time an \texttt{Operator} reads or writes to (part of) a \texttt{Signal}, the \texttt{SignalDict} will direct that information to the appropriate \texttt{Variable}.\footnote{Note that although we are using \texttt{Variables} for all the \texttt{Signals}, not all \texttt{Signals} are trainable; we still only optimize the \texttt{Signals} corresponding to trainable parameters of the model (e.g., connection weights and biases).}  We still need the \texttt{SignalDict} bookkeeping because we need to make sure that reads and writes to the \texttt{Variable} happen in the right order.  So, for example, when an \texttt{Operator} reads from a \texttt{Variable} $v$ it reads from the version of that variable \emph{after} any other \texttt{Operators} have made their updates.  The \texttt{SignalDict} keeps track of those versions, and directs the reads and writes to the appropriate place.

\subsubsection{Operators}

With this infrastructure in place, the mapping from Nengo \texttt{Operators} to TensorFlow \texttt{Ops} is relatively straightforward.  Every \texttt{Operator} implementation follows the same basic structure:

\begin{enumerate}
\item Get the current value of all the input \texttt{Signals} from the \texttt{SignalDict}
\item Apply TensorFlow ops that implement the desired computation
\item Use the \texttt{SignalDict} to write the results back to any output \texttt{Signals}
\end{enumerate}

Thus we can create a small computational subgraph consisting of reads, transformations, and writes that will implement each Nengo \texttt{Operator}.  The subgraphs for different \texttt{Operators} are connected via the \texttt{Signals} (represented as \texttt{Variables}) that they read and write.  So as we iterate through all the Nengo \texttt{Operators} and add them into the TensorFlow graph, we gradually build up a complete graph of interconnected \texttt{Ops} that will implement a single timestep of a Nengo simulation.

The final step is to embed this single timestep within a framework that will simulate the model over time.  For this we can use TensorFlow's \texttt{tf.while\_loop}, which is a way to represent a loop using TensorFlow's declarative programming style.  Generally speaking this will meet all of our needs, although some bookkeeping is needed to make sure that computations from different timesteps do not overlap incorrectly.  The only concern is that \texttt{tf.while\_loop} adds a certain amount of overhead to every iteration, which can slow down the simulation.  Thus NengoDL has an option to unroll the simulation loop by explicitly building multiple timesteps into the TensorFlow computation graph.  Essentially we go through the same process as above to build a single timestep, then repeat that $n$ times so that we end up with $n$ implementations of every Nengo \texttt{Operator} (all connected together in the correct order thanks to the \texttt{SignalDict}).  We then embed that whole thing within a \texttt{tf.while\_loop}, so that every iteration will execute $n$ timesteps.  This results in a more complicated TensorFlow graph, which increases the build time and memory usage, but can significantly improve the simulation speed.  This functionality is accessed through the \texttt{unroll\_simulation} parameter of \texttt{nengo\_dl.Simulator}, where \texttt{nengo\_dl.Simulator(net, unroll\_simulation=10)} indicates that we should unroll the simulation as above with $n=10$.

\subsection{Graph optimizations}
\label{sec:graphoptimizations}

Naively implementing the above process results in a functional simulator, but a slow one.  The core problem is that every time an \texttt{Op} is executed TensorFlow has to launch a kernel, and there is a certain amount of associated overhead (especially when launching kernels on the GPU).  If we have many small kernel launches, any benefits of the underlying accelerator will be lost in that overhead.  So when building an efficient neural simulator in TensorFlow it is important that we try to combine operations as much as possible, so that we end up with fewer, larger kernels.  For example, imagine we have 10 \texttt{ElementwiseInc} operations, each reading two signals and multiplying them together.  Implemented individually, this would be 20 reads, 10 multiplies, and 10 writes.  It would be much better to combine those operations together into one large op that would do two reads, one multiply, and one write.  NengoDL automatically applies a number of these kinds of optimizations to the \texttt{Operator} graph in order to improve performance.  All of these optimizations are transparent to the end user, neither requiring their input nor modifying the model's output (other than making it faster).

\subsubsection{Merging}
\label{sec:merging}

\begin{figure}
\centering
\includegraphics[width=0.9\textwidth]{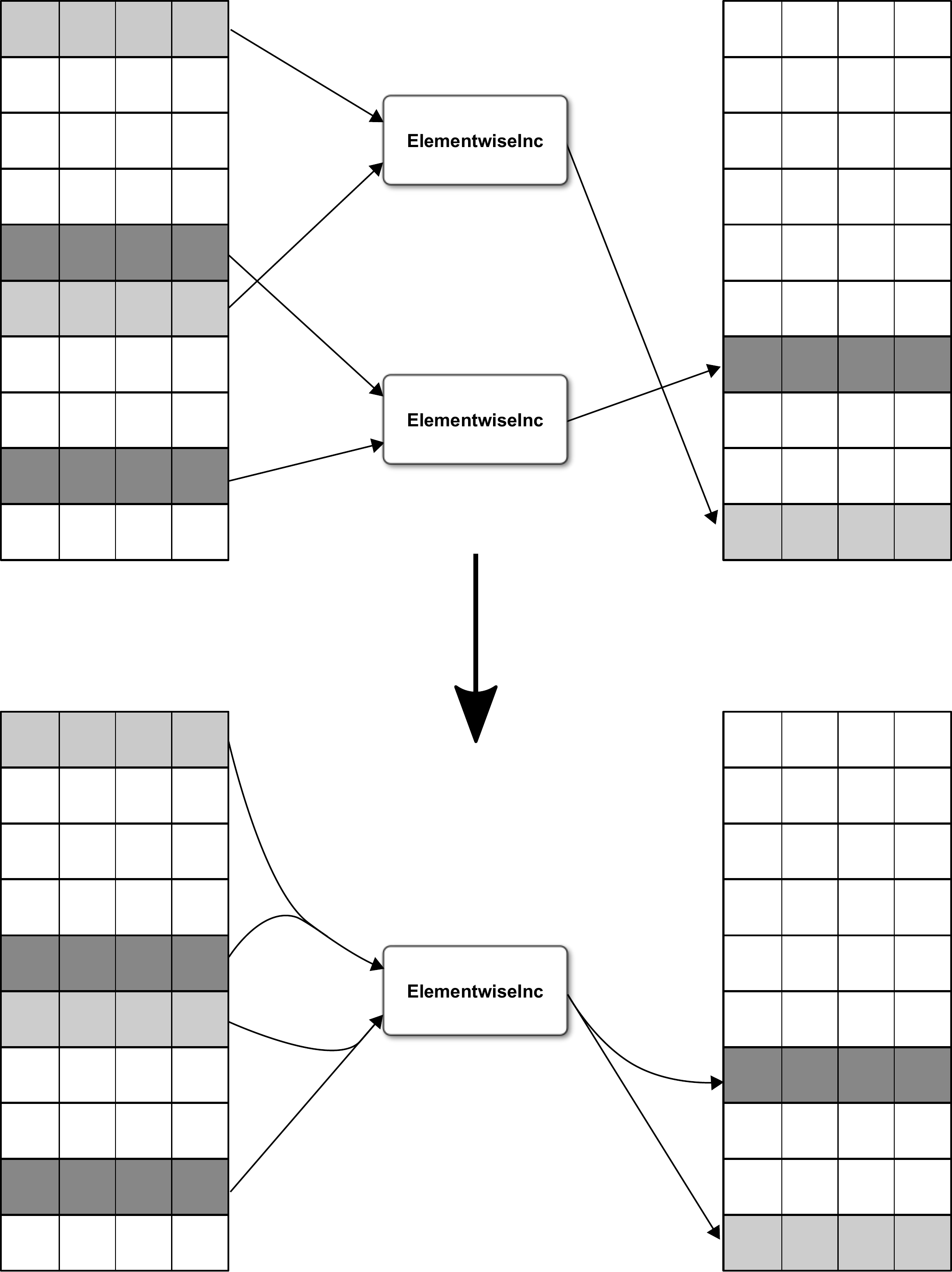}
\caption{Illustration of operator merging.  Signals have been merged into combined base arrays.  We begin with two \texttt{ElementwiseInc} operators that each read two input signals (subsets within those base arrays), multiply them together, and write the result to some output signal.  To merge the operators we combine the reads, do a single multiply, and write the combined result.}
\label{fig:merging}
\end{figure}

The first step is to merge \texttt{Signals}, so that we can read and write larger chunks of data.  We do this by concatenating them along the first dimension, e.g. combining two $10 \times 5$ arrays into one $20 \times 5$ array.  Note that this requires that the array shapes match on all dimensions beyond the first (i.e., we could not merge a $10 \times 5$ with a $10 \times 6$ array).  The arrays also need to have the same type (e.g., integer versus float) and other TensorFlow meta information, such as trainability.  Thus we will still end up with various different base arrays, but a much smaller number than we started with.

To track these new data structures NengoDL defines a new object called a \texttt{TensorSignal}, which stores a reference to a base array and some indices.  We then translate every \texttt{Signal} into a \texttt{TensorSignal}, which indicates where the data for that \texttt{Signal} is stored and which elements in that base array (indexed along the first axis) contain the data for that \texttt{Signal}.  So whereas before an \texttt{Operator} would read/write to some \texttt{Signal}, instead it will read/write to some subset of the base array, as specified by the corresponding \texttt{TensorSignal}.  Merging multiple reads into one is then as easy as combining their indices (as long as all the reads have the same base array).

The next step is to merge the operations themselves (e.g., combining the ten multiplies into one).  Generally speaking, two operations are mergeable if each of their inputs and outputs are mergeable (have the same base array).  For example in the \texttt{ElementwiseInc} case, once we are able to read each input as one large chunk, we can do a single \texttt{tf.multiply} to multiply them all together at once (Figure~\ref{fig:merging}).  There are some additional caveats when merging more complex operators, which depend on the details of those operators, but we will not go into that here.

The other main concern for merging operators is that we cannot merge two operators if the input to one depends on the output of the other.  This would introduce a circular dependency if we tried to compute those two operations simultaneously.  Fortunately, Nengo already organizes all the \texttt{Operators} into a dependency graph, in order to schedule their execution (e.g., so that reads and writes to a \texttt{Signal} happen in the correct order).  So we can use that graph to determine whether or not two operators depend on each other, and therefore whether or not they are mergeable.

\subsubsection{Planning}
\label{sec:planning}

\begin{figure}
\centering
\includegraphics[width=0.9\textwidth]{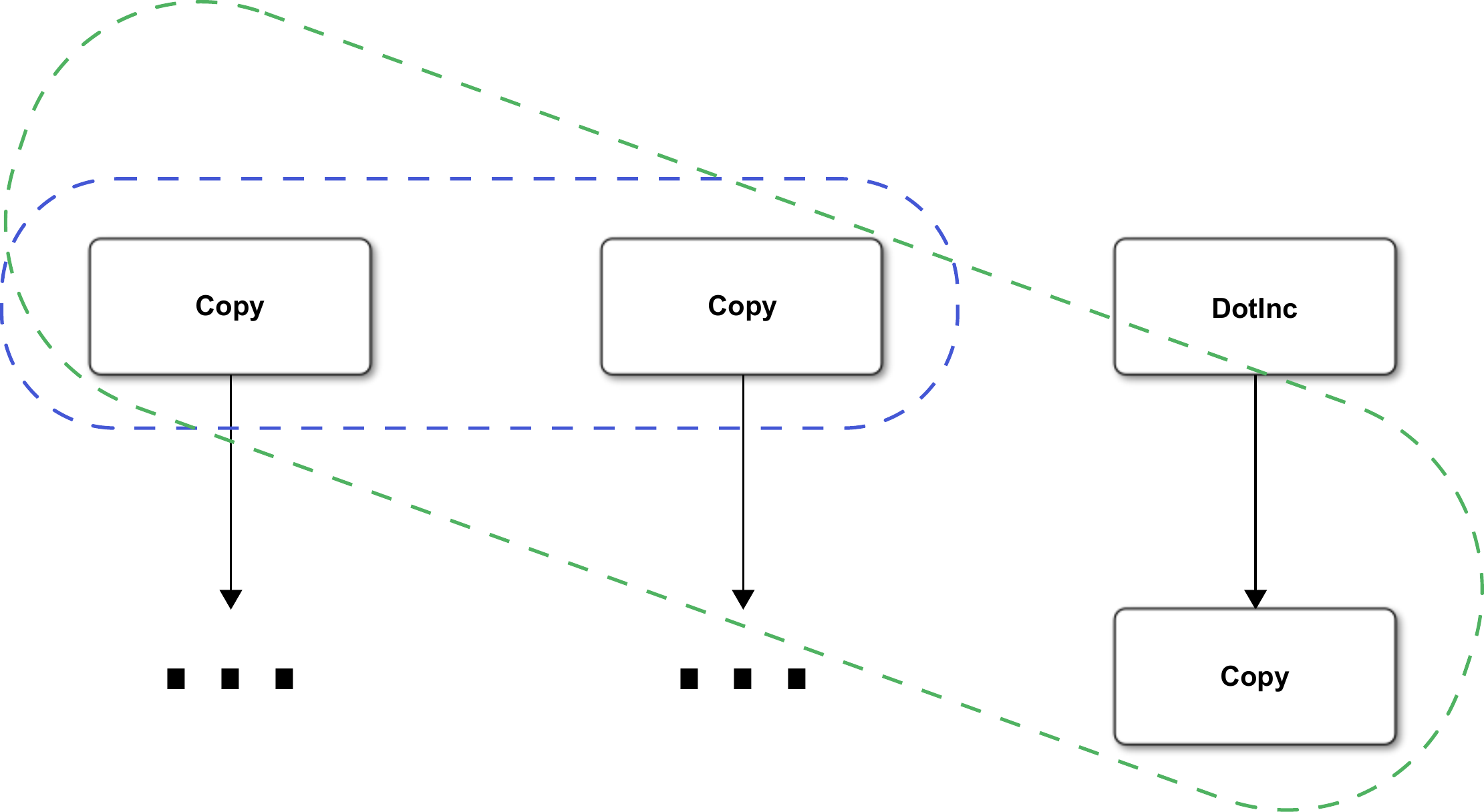}
\caption{Example of operator execution order planning.  Arrows indicate signal read/write dependencies.  By scheduling the \texttt{DotInc} operator first, we are able to more efficiently schedule the three \texttt{Copy} operators as a single group.}
\label{fig:planning}
\end{figure}

When optimizing operator merging we also need to consider the order in which (groups of) \texttt{Operators} are executed, because the execution order can affect which operators can be merged.  We can see an example of this in Figure~\ref{fig:planning}.  At first two \texttt{Copy} operators and one \texttt{DotInc} operator are available to be scheduled (as they have no incoming dependencies).  One might be tempted to schedule the two \texttt{Copy} operators first (blue circle), as that allows us to combine the two \texttt{Copy} ops into one.  However, if we schedule the \texttt{DotInc} operator first then the third \texttt{Copy} operator will be freed of its dependency, and we will be able to merge all three \texttt{Copy} operators (green circle).  This is a simple example, but the problem rapidly becomes much more complex, and efficient merging becomes more important, as the number of operators increases.  Thus the goal of the planning process is to take an (unordered) list of \texttt{Operators}, and try to find an order of execution that best promotes the efficient merging of operators.

NengoDL includes a number of different planning methods, such as a greedy algorithm that simply selects the largest available group of operators to be scheduled next, or a method based on analyzing the transitive closure of the dependency graph (with some heuristic prioritization), inspired by \citet{Gosmann2017}.  However, the method that we found to provide the best tradeoff between plan quality and optimization time, in general, is a bounded breadth-first tree search.  That is, we search through all possible execution plans up to some length $n$, and then schedule the group of operators corresponding to the first step in the best plan we find (``best'' defined as the plan that schedules the most total operators in $n$ steps).  We then repeat this process on the remaining operators, until all operators have been scheduled.  For $n=1$ this corresponds to the greedy algorithm, and for $n=\infty$ we find the optimal plan (the plan with the shortest number of steps).  A reasonable value for $n$ depends on the complexity of the model and the available computational budget; however, we find that $n \approx 3$ works well in practice.

\subsubsection{Sorting}

\begin{figure}
\centering
\includegraphics[width=0.9\textwidth]{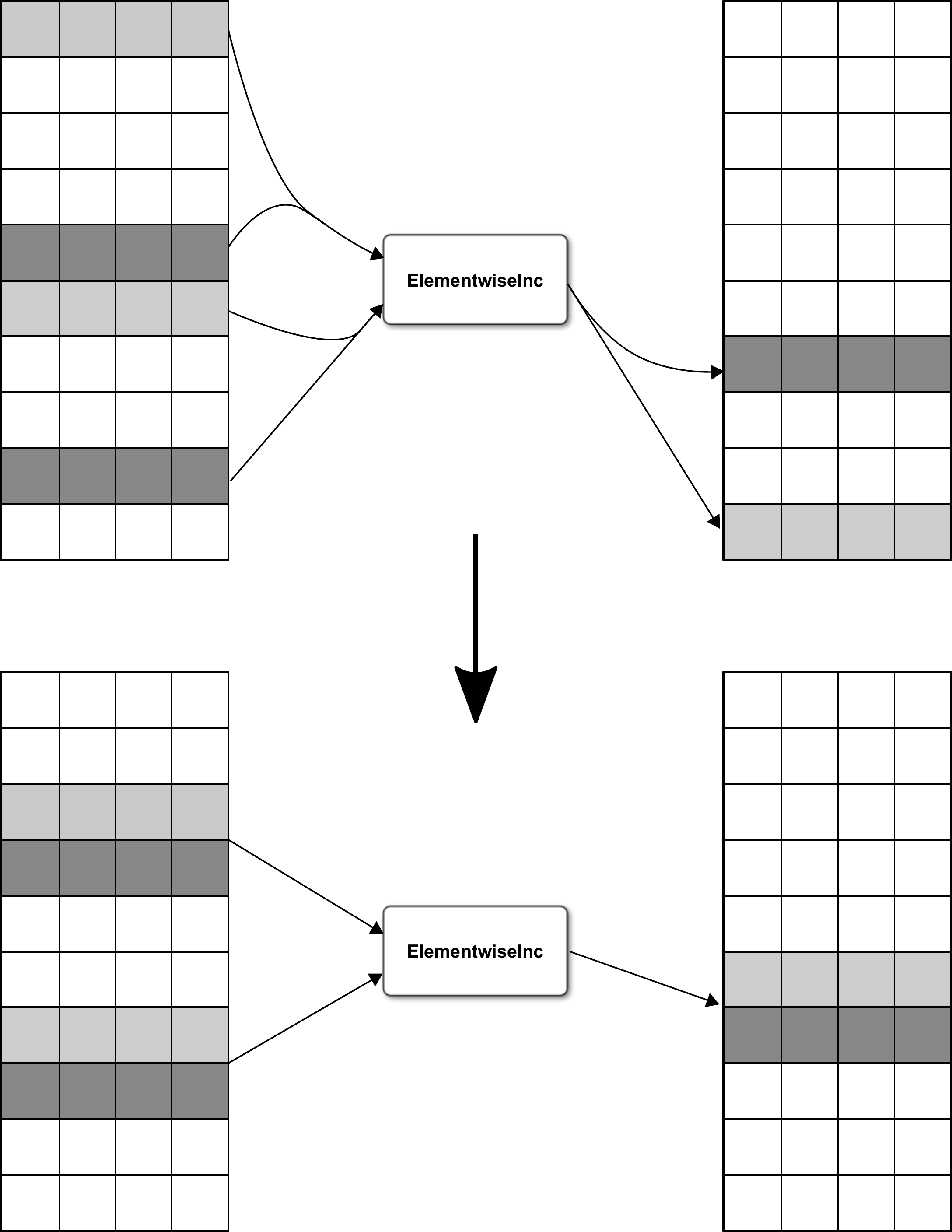}
\caption{Illustration of signal sorting (continuing the example from Figure~\ref{fig:merging}).  By rearranging the signals into ordered, contiguous blocks we can increase the efficiency of the read operations.}
\label{fig:sorting}
\end{figure}

Another important optimization concern is the order in which the \texttt{Signals} are arranged in memory.  Recall that \texttt{Signals} are combined into large TensorFlow \texttt{Variables}, and when we read from a \texttt{Signal} we are reading from some subset of indices within that \texttt{Variable}.  However, it is faster to read from a contiguous, in-order block of indices (e.g. 5, 6, 7, 8), rather than an arbitrary set of indices (e.g., 5, 10, 12, 20, or 5, 8, 7, 6).  So we want to try to arrange the \texttt{Signals} within the \texttt{Variables} such that \texttt{Signals} that are read by the same group of \texttt{Operators} are adjacent and in the same order as the \texttt{Operators} (Figure~\ref{fig:sorting}).

The challenge is that we have many different groups of \texttt{Operators}, reading from possibly overlapping sets of \texttt{Signals}, such that reordering signals with respect to one group of \texttt{Operators} may break the contiguity with respect to a different set of \texttt{Operators}.  We also need to consider the order of the \texttt{Operators} within a group; we can rearrange the \texttt{Operators} to promote efficient reads, rather than reordering the \texttt{Signals}.  For example, if the \texttt{Signals} are arranged in the order 4, 1, 2, 3, but we move the first \texttt{Operator} in the group (which reads from the \texttt{Signal} with index 4) to the end, then this becomes an in-order, efficient read.  The reason why we might want to rearrange \texttt{Operators}, rather than just changing the order of the \texttt{Signals}, is that the 4, 1, 2, 3 order may be an efficient order for a different group of \texttt{Operators} reading from an overlapping set of \texttt{Signals}.  Yet another issue is that a single group of \texttt{Operators} can be reading from multiple blocks of \texttt{Signals}, meaning that if we change the order of the \texttt{Operators} we change the order of the reads within all of those \texttt{Signal} blocks (possibly changing some other block that used to be in-order to now be out-of-order).

We end up with a complex constraint satisfaction problem, where we are trying to find the \texttt{Signal}/\texttt{Operator} ordering that will result in the best possible read performance. A perfect solution, where every read is a contiguous block, is usually not possible, nor is there an efficient algorithm for finding an optimal solution (that we know of).  We arrived at a solution that uses some heuristic prioritization and an iterative settling procedure to try to find an ordering that works well in practice.  

The first step is to sort the \texttt{Signals} into contiguous blocks, without worrying about order.  The \texttt{Operators} are already arranged into groups by the planning process described above, so we know all the \texttt{Signals} that will be read by each group of \texttt{Operators} (which we will call a read block).  We can then group all the \texttt{Signals} according to which set of read blocks they participate in, which we will call a meta-block.  If all the read blocks were non-overlapping, then every meta-block would contain a single read block; however, this is rarely the case.  Since the order of \texttt{Signals} within a meta-block does not matter (yet), we can reformulate the problem as sorting the meta-blocks into an order that ensures the underlying read blocks are as contiguous as possible.  Again, an ideal sorting is usually not possible, so in general we prioritize the contiguity of larger blocks (as they are the more expensive read operations).

\begin{algorithm}
\DontPrintSemicolon
initialize list of all meta-blocks $M$\;
initialize sorted list $S$ (empty)\;
set active read block $a$ to be the largest read block\;
set active meta-block $c$ to be any meta-block containing $a$\;
\While{$M$ is not empty}{
	$X \leftarrow \{m \in M \mid a \in m\}$\;
	\uIf{$X = \emptyset$}{
		$X \leftarrow M$\;
		$a \leftarrow$ largest read block in $c$\;
	}
	$Y \leftarrow \{x \in X \mid c \subseteq x\}$\;
	\uIf{$Y = \emptyset$}{
		$Y \leftarrow X$\;
	}
	$Z \leftarrow \{y \in Y \mid | y \oplus c | = \min_{n \in Y} | n \oplus c |\}$\;
	\While{|Z| > 1}{
	  $m \leftarrow$ the next largest read block in $c$\;
		$Z \leftarrow \{z \in Z \mid m \in z\}$\;
	}
	remove the remaining $z \in Z$ from $M$ and append it to $S$\;
	$c \leftarrow z$\;
}
\caption{Meta-block sorting algorithm}
\label{alg:metablocksort}
\end{algorithm}

Our meta-block sorting algorithm is shown in Algorithm~\ref{alg:metablocksort}.  Broadly speaking, this algorithm tries to find the next meta-block that best matches the last meta block we selected.  Matching is determined by narrowing down the set of remaining meta-blocks according to increasingly strict criteria: 1) any meta-blocks that contain the active read block (so that we know that at least the active block will end up being contiguous); 2) any meta-blocks that contain all the elements in the last meta-block; 3) the meta-blocks with minimal Hamming distance to the last meta-block; and 4) the meta-block that contains the largest read blocks in the last meta-block.

After the meta-block sorting process, the \texttt{Signals} are arranged into (semi-) contiguous blocks of indices within the base \texttt{Variables}.  We then want to sort the \texttt{Operators} and \texttt{Signals} within each meta-block so that the indices are in increasing order.  Recall that because our \texttt{Signal} blocks overlap, and because a group of \texttt{Operators} can read from multiple \texttt{Signal} blocks, there is unlikely to be an ordering that satisfies all the constraints.  Again we prioritize larger read blocks.

\begin{algorithm}
\DontPrintSemicolon
sort the list of read blocks $B$ by increasing order of size\;
\For{$i = 1 \rightarrow n$}{
	\For{$b \in B$}{
		$O \leftarrow$ the set of \texttt{Operators} associated with $b$\;
		sort $O$ to match the order of the \texttt{Signals} in $b$\;
		$C \leftarrow$ the set of read blocks associated with $O$\;
		\For{$c \in C$}{
			sort the signals in $c$ to match the order of $O$\;
		}
	}
	if the order of the \texttt{Signals}/\texttt{Operators} did not change, terminate early\;
}
\caption{Signal/Operator sorting algorithm}
\label{alg:sigopsort}
\end{algorithm}

Algorithm~\ref{alg:sigopsort} cycles between two steps: 1) sort the \texttt{Operators} to match the order of a given \texttt{Signal} block $b$, and 2) sort all the \texttt{Signal} blocks read by that group of \texttt{Operators} to match the new order of the \texttt{Operators} (this sorting is restricted such that it cannot change the meta-block order).  The basic idea is that after step 1, we know that $b$ will be contiguous and in-order (assuming that the meta-block sorting algorithm was able to make $b$ contiguous).  However, imagine that our \texttt{Operator} group also reads from another block $c$.  Rearranging the order of the \texttt{Operators} may have put $c$ out of order, so we fix that in step 2.  

Note, however, that there may be some other group of \texttt{Operators} that also reads from $b$ or $c$ (or some overlapping set).  Thus the sorting we just performed might have broken an earlier ordering we established for that other group of \texttt{Operators}.  That is why we iterate over the read blocks in increasing order of size; we know that later sorting will only break the ordering of earlier, and therefore smaller, blocks.  However, it is possible that after the \texttt{Signals} are reordered by a larger block (step 2), the \texttt{Operators} in a smaller block could be reordered to match that new \texttt{Signal} order (step 1).  That is why we perform multiple passes over the read blocks, to allow the smaller blocks to settle into orderings that are consistent with the larger blocks.

\subsubsection{Simplification}

Another optimization we perform is to simplify the \texttt{Operator} graph by checking for certain special case combinations of \texttt{Operators}.  For example, we can change $y\mathrel{+}=x*1$ to $y\mathrel{+}=x$ in order to save a multiplication, or if there is a \texttt{Copy} operation that moves data from $x$ to $y$, but the value of $x$ never changes, we can change that to a \texttt{Reset} operator that directly sets the value of $y$ to that constant value (saving a read).  These optimizations do not have a large impact relative to the merging and sorting, but they are also relatively simple and quick to perform.

\section{Results}
\label{sec:results}

There are two areas we will focus on in the results: the simulation speed of NengoDL, and some practical demonstrations of using NengoDL to construct and optimize a neural model.  The code needed to reproduce any of the results presented here is available at \url{https://github.com/nengo/nengo-dl/tree/master/docs/whitepaper}. 

\subsection{Simulation speed}
\label{sec:simulation_speed}

\begin{figure}
\centering
\includegraphics[width=\textwidth]{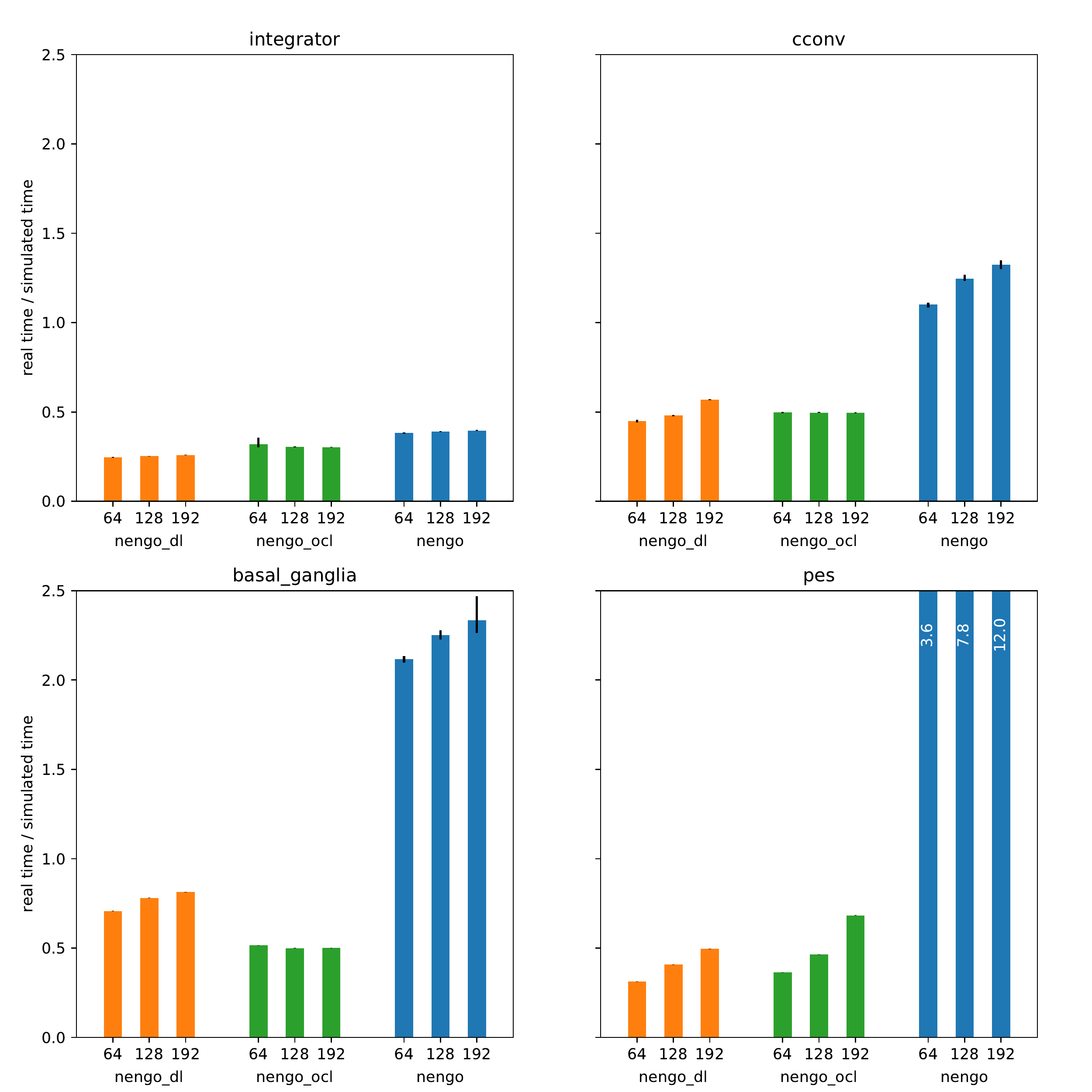}
\caption{Comparing simulation speed of NengoDL versus NengoOCL versus Nengo on various benchmark models.  Error bars indicate 95\% confidence intervals on the mean over 5 runs.  We show scaling with respect to the represented dimensionality (64, 128, 192).}
\label{fig:compare_backends_1}
\end{figure}

We compare the simulation speed of NengoDL to the default Nengo simulator (which is CPU only) as well as NengoOCL (a simulator that runs on the GPU using custom OpenCL kernels).  All results are collected using an Intel Xeon E5-1650 3.5GHz CPU and an Nvidia GeForce GTX Titan X GPU (in the case of NengoDL and NengoOCL).

Figure~\ref{fig:compare_backends_1} shows the relative speed of the simulators on four different benchmark models.  The purpose of the models is not particularly important; they were simply chosen to showcase a range of different models with varying complexities:

\begin{itemize}
\item {\bf integrator}: A single ensemble of recurrently connected neurons (a common component used to implement a memory system in Nengo)
\item {\bf cconv}: A network implementing the circular convolution of two input vectors (commonly used in Nengo Semantic Pointer Architecture models)
\item {\bf basal\_ganglia}: A model of basal ganglia circuitry, commonly used to perform action selection
\item {\bf pes}: An ensemble of neurons with output weights that are updated as the simulation runs, using the Prescribed Error Sensitivity learning rule \citep{MacNeil2011}
\end{itemize}

In order to get a better picture how the different backends compare, we show how the performance scales as we change a parameter of these model, the represented dimensionality. This increases the complexity of the model in a number of ways; for example, it increases the number of neuron ensembles, the dimensionality of the signals being transmitted throughout the model, and the number of parameters (through the size of encoder/decoder matrices).

Overall we can see that the GPU-based simulators (NengoDL and NengoOCL) offer significant performance improvements, with NengoOCL or NengoDL offering the best performance on different benchmarks.

\begin{figure}
\centering
\includegraphics[width=\textwidth]{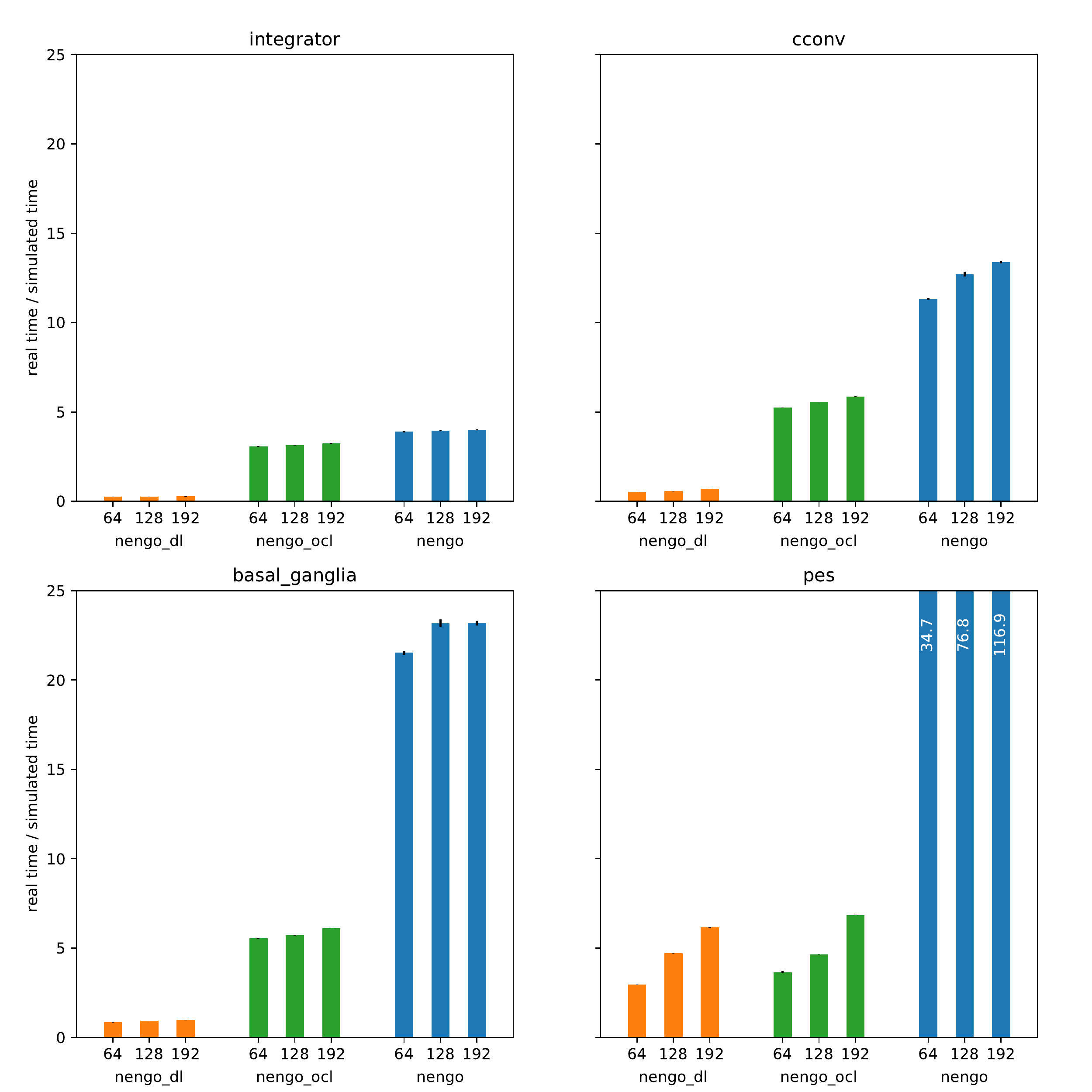}
\caption{Comparing simulation speed of NengoDL versus NengoOCL versus Nengo on various benchmark models with 10 batched inputs.  We show scaling with respect to the represented dimensionality (64, 128, 192).  Note that we do not get the NengoDL batching benefits on the PES benchmark, because that network applies an online learning rule to the weights (meaning that we need a separate weight matrix for each batch element).}
\label{fig:compare_backends_10}
\end{figure}

That being said, we can see an important advantage of NengoDL in Figure~\ref{fig:compare_backends_10}.  In this case we are running the same benchmarks, but we are running each model ten times.  With Nengo and NengoOCL this involves serially running the model ten times in a row, which, unsurprisingly, takes about ten times as long.  However, NengoDL allows models to be run with batched inputs, so we can simulate the model once with ten different inputs in parallel.  This scales much better as we increase the batch size, thanks to the parallelism of the computations.  Thus if a modeller wants to test their model with a range of different inputs, NengoDL will probably offer the best performance.

\begin{figure}
\centering
\includegraphics[width=0.8\textwidth]{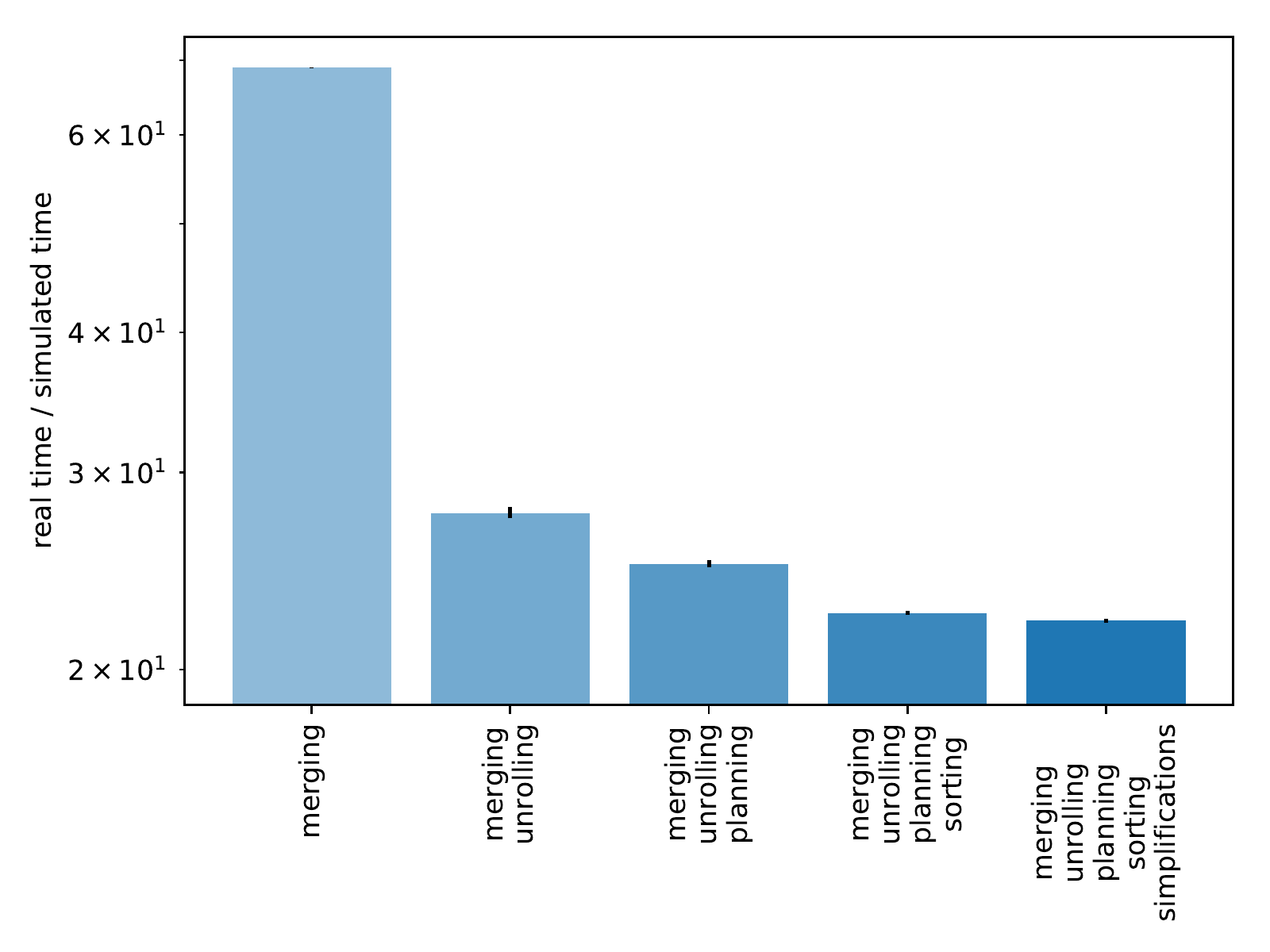}
\caption{Impact of the various NengoDL graph optimization methods on the simulation speed of the Spaun model.  Note that the speed is being plotted on a logarithmic scale.  {\bf Merging}: multiple operators of the same type are combined into a single, larger operator of that type (using a greedy planner).  Without this step the simulation speed is extremely slow, so we include it in all cases.  {\bf Unrolling}: the simulation loop is unrolled within the TensorFlow computation graph.  {\bf Planning}: A more advanced planning algorithm is used (the tree planner) to promote better operator merging.  {\bf Sorting}: Signals are sorted to promote more efficient reads.  {\bf Simplifications}: Unnecessary \texttt{Operations} are removed from the Nengo build graph.  More details on all the optimization methods can be found in Section~\ref{sec:graphoptimizations}.}
\label{fig:compare_optimizations}
\end{figure}

Finally, it is interesting to explore the effect of the various graph optimization steps described in Section~\ref{sec:graphoptimizations}.  Figure~\ref{fig:compare_optimizations} shows the speed of NengoDL when simulating the Spaun model (an updated version of \citealt{Eliasmith2012a}, available at \url{https://github.com/xchoo/spaun2.0}) with 128-dimensional vectors, consisting of 1.2 million neurons split amongst 21k ensembles with 91k connections.  Spaun was chosen because the complexity of this model provides a good stress test for the graph optimization methods.  We can see that each type of optimization provides incremental improvements to the simulation speed.  Note that in the ``planning'' case we are comparing the tree planner to the greedy planner (see Section~\ref{sec:planning}), rather than the presence and absence of planning.  That is, in all cases we are performing operator merging.  If we do not perform any merging then the simulation is extremely slow (after one hour the simulation had still not finished initializing the TensorFlow graph).

\subsection{Model examples}

Simulation speed is an important aspect of NengoDL, but equally important are the novel features NengoDL provides that are not available in any other Nengo simulator.  Specifically, NengoDL includes the ability to: a) insert TensorFlow components, such as convolutional layers, into a Nengo model; b) convert rate-based deep learning networks into spiking versions; and c) optimize the parameters of a Nengo model using deep learning training methods.  In this section we will present some basic examples illustrating these features and the advantages they provide.

\subsubsection{Spiking MNIST}
\label{sec:spiking_mnist}

In this model we use the \texttt{TensorNode}/\texttt{tensor\_layer} syntax to create a simple convolutional network in Nengo, consisting of three convolutional layers, two average pooling layers, and a dense linear readout.  We use the Leaky Integrate and Fire neuron model, which has both a rate and spike-based implementation.  As described in Section~\ref{sec:training}, we use NengoDL to automatically swap between the differentiable rate implementation during training and the spiking model during testing/inference.  We also take advantage of NengoDL's ability to smoothly combine TensorFlow and Nengo models; we use \texttt{TensorNodes} to implement convolutional and pooling layers using TensorFlow, combined with standard Nengo \texttt{Ensembles} to implement the neural nonlinearities and \texttt{Connections} to link layers together.  

We train the model on the deep learning ``hello world'' task, MNIST digit classification (the model receives an image of a hand-written digit as input and must classify that digit 0--9).  This kind of vision system has been integrated with cognitive models in, e.g., \citet{Eliasmith2012a}, where the model used an MNIST vision system combined with working memory, inductive reasoning, and motor control capabilities to perform a range of different cognitive tasks.  However, in that case the vision system was trained separately using a standard deep learning package, and then imported into Nengo.  Here we show that using the new features of NengoDL we can directly build and train these deep learning style networks within the Nengo framework, making it much simpler to construct integrated, hybrid cognitive model as in \citet{Eliasmith2012a}.

After training, the model achieves 99.05--99.09\% classification accuracy (95\% confidence intervals), which is the performance we would expect for MNIST.  However, one of the important features of Nengo, which is retained in NengoDL, is the ability to smoothly switch between rate and spiking neuron models.  After training the model using the rate-based implementation of LIF neurons, we can then run the model using spiking LIF neurons (using the same trained weights).  This spiking version of the network achieves 98.41--98.87\% classification accuracy, only a small decrease from the rate version.  Spiking deep learning is an interesting and active research field \citep{Hunsberger2016,Lee2016}, and one which NengoDL is naturally situated to support.

\subsubsection{Memory storage and retrieval}

In the second example we want to explore the application of the NengoDL training functionality to a more cognitive/neuromorphic style of model, rather than a standard deep learning vision network.  We construct a model using Nengo's Semantic Pointer Architecture (SPA), which uses high-dimensional vectors, encoded in neural activity, to represent structured symbolic information.  We apply the model to a memory retrieval task: the network is given a sequence of attribute-value pairs as input (e.g. $(colour: red)$, $(shape: circle)$, $(texture: smooth)$) that it must dynamically store in memory using neural activities.  At a later point the network is prompted with one of the attributes (e.g., $shape$), and must respond with the corresponding value (e.g., $circle$).  Note that we want to perform this task for arbitrary input sequences, so the solution cannot be directly built into the connection weights (i.e., we cannot just train the network to output $circle$ when it gets the cue $shape$).  We want the network to learn the abstract functions required to store and retrieve generic items from memory.

The network architecture consists of a circular convolution network (to combine the attribute-value pairs into a linked representation), a recurrently connected ensemble of neurons to implement the memory, and a final circular convolution network to extract the cued attribute from the memory.  In this example we use rectified linear neurons.  We can construct this model, without NengoDL, by using the  least-squares-based optimization methods standard in Nengo.  The advantage of these methods is that they are fast and do not rely on gradient descent (and therefore do not require the model to be differentiable).  However, these methods can only optimize the output weights of one ensemble/layer at a time.  This means that each layer in the above model is optimized independently, and there are many parameters (e.g., input weights, gains, and biases) that are not optimized (they are typically chosen from some random distribution).  By using NengoDL we can begin with the standard Nengo optimized model, and then apply the deep learning optimization on top of that.  This allows us to jointly optimize across the layers of the model, and fine tune all the parameters in the model as well as the output weights.

We train the model by generating a randomized set of training data, with different sequences of attribute-value pairs and different vector vocabularies.  For each of these inputs we can specify what the correct model output would be.  We can then use TensorFlow's gradient-descent based optimizers (in this case, RMSProp; \citealt{Tieleman2012}) to optimize all the parameters of the model with respect to those inputs and target outputs.  The details of the training hyperparameters can be found in the code at \url{https://github.com/nengo/nengo-dl/tree/master/docs/whitepaper}.

\begin{figure}
\centering
\includegraphics[width=0.8\textwidth]{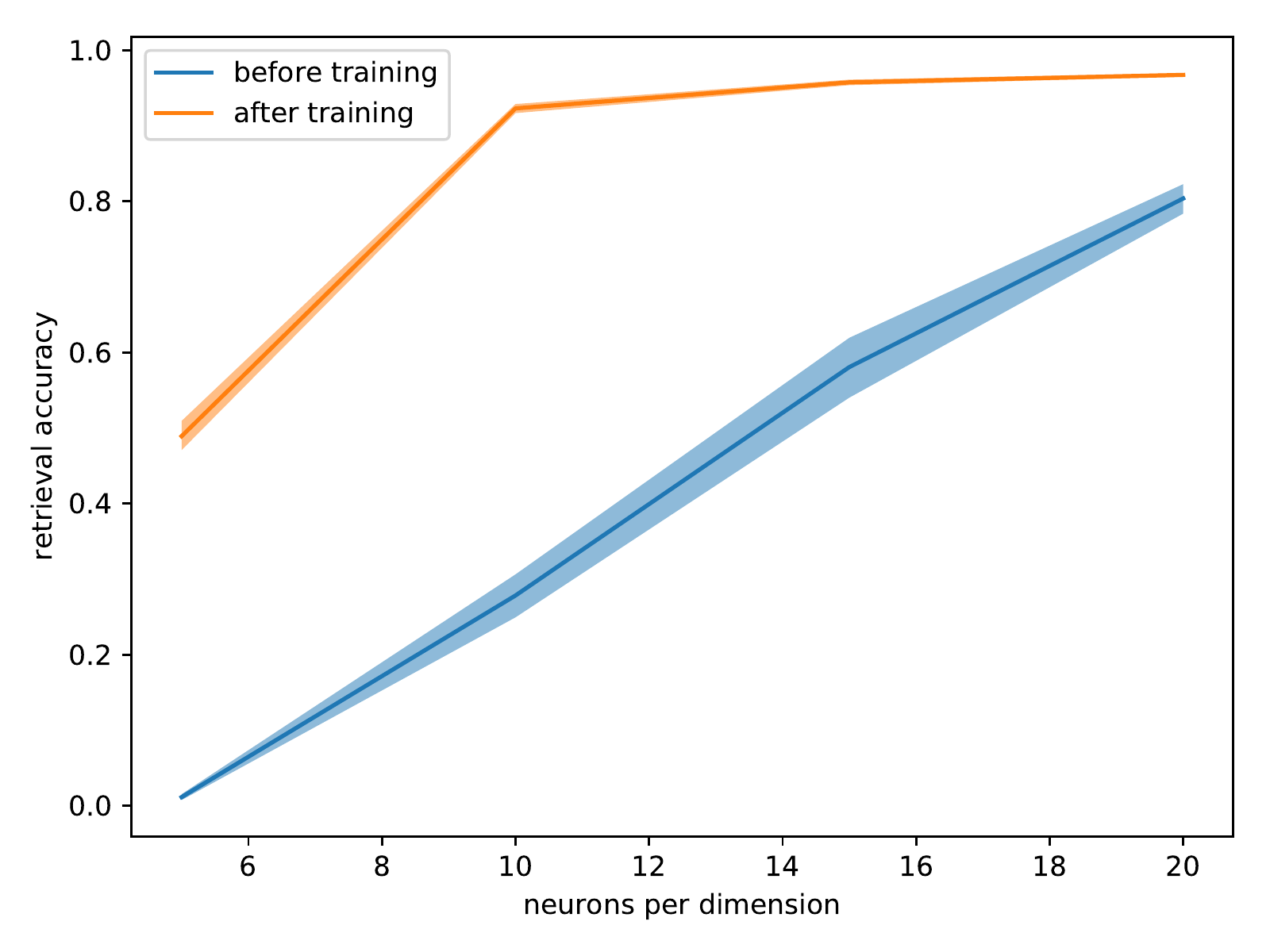}
\caption{Retrieval accuracy on the memory task, before and after training is applied.  Showing 95\% confidence intervals (for different random initializations).}
\label{fig:spa_optimization}
\end{figure}

The effect of the training is shown in Figure~\ref{fig:spa_optimization}.  This figure shows the retrieval accuracy of the model (on a separate set of randomly generated test data), which is computed by comparing the output of the model (which should be the value of the cued attribute) to all the vectors in the vocabulary (e.g., $colour$, $shape$, $red$, $circle$, etc.).  If the output of the model is most similar to the correct answer, then that is a successful retrieval.  We can see that the performance of the model is significantly improved after applying the NengoDL training.  In particular we can see the impact of the training for smaller numbers of neurons.  This makes sense given the random initialization of many of the neural parameters under the standard Nengo methods. For larger numbers of neurons that random initialization is likely to give a good-enough coverage of the parameter space, but for smaller numbers it is more important that those parameters be fine-tuned for the problem.  In other words, one important advantage of the NengoDL optimization features are that they allow us to take better advantage of limited neural resources.

\section{Conclusion}

The goal of NengoDL is to provide a tool that unites deep learning and neuromorphic modelling methods.  It combines the robust modelling API of Nengo with the speed and optimization methods of TensorFlow.  This allows the modeller to build complex cognitive/neuromorphic models, combine them with deep learning elements (such as convolutional layers), simulate them efficiently, and optimize their parameters using modern deep learning training methods.

In this paper we have introduced the features and some interesting implementation aspects of NengoDL.  Those interested in learning more or using NengoDL in their own work can find much more information in the online documentation at \url{https://www.nengo.ai/nengo-dl}.  This includes installation instructions, details on all the novel features of NengoDL and how to access them, as well as examples illustrating various different styles of models.  All the source code can be found at \url{https://github.com/nengo/nengo-dl}.  There is also a forum at \url{https://forum.nengo.ai} where users can get help with specific questions.  Finally, NengoDL is under active development; feel free to suggest features on the forum or at \url{https://github.com/nengo/nengo-dl/issues} so that we can continue to improve this tool for the modelling community.

\subsubsection*{Acknowledgments}

This work was supported by Applied Brain Research, Inc. and ONR MURI N00014-16-1-2832.

\small

\bibliographystyle{abbrvnat}

\end{document}